\documentclass[extendedabs]{recpad2k}

\usepackage{adjustbox}
\usepackage{array}
\usepackage{booktabs}
\usepackage{multirow}
\usepackage{graphicx}
\usepackage{slashbox}

\title{Action Recognition for American Sign Language}

\addauthor{Nguyen Huu Phong}{phong@dei.uc.pt}{1}
\addauthor{Bernardete Ribeiro}{bribeiro@dei.uc.pt}{1}
\addinstitution{
CISUC -- Department of Informatics Engineering\\
University of Coimbra, Polo II, Pinhal de Marrocos, \\3030--290 Coimbra, Portugal\\
}

\runninghead{Student, Prof, Collaborator}{RECPAD Author Guidelines}

\begin{document}
\maketitle

\begin{abstract}
In this research, we present our findings to recognize American Sign Language from series of hand gestures. While most researches in literature focus only on static handshapes, our work target dynamic hand gestures. Since dynamic signs dataset are very few, we collect an initial dataset of 150 videos for 10 signs and an extension of 225 videos for 15 signs. We apply transfer learning models in combination with deep neural networks and background subtraction for videos in different temporal settings. Our primarily results show that we can get an accuracy of $0.86$ and $0.71$ using DenseNet201, LSTM with video sequence of 12 frames accordingly.
\end{abstract}
\section{Introduction}
\label{sec:intro}
The use of deep learning (DL) for action recognition in video has been an active research in recent years. However, they often fall into recognition of actions as the whole picture e.g. an actor is running, jogging or walking or activities of a group playing football, tennis so on and so forth. Recognizing a more subtle action can be seen in emotion or a more relevant to our research as hand gesture. Though hand gesture is often be limited by number of actions e.g. moving left and right, pointing or twisting etc. On the contrary, sign language offers a more standard and abundance of vocabularies for actions. Just only in America, there are approximately a half of million people using American Sign Language (ASL). 

Research in sign languages mostly focus on static images e.g. numbers or alphabets. For example in ASL alphabets, letters J and Z which are dynamic signs are excluded \cite{kuznetsova2013real,ameen2017convolutional}. Some works explore continuous signs -- shown as continuous frames -- but vocabularies are just static signs \cite{koller2016deep}. In our research, we analyze dynamic ASL signs where a sign requires at least two or more shapes.

The structure of this paper is as follows. In the next section, we discuss about datasets gathered for this research and present our framework in Section 3. In Section 4, experiments are  demonstrated and results analyzed. Conclusions and future works are addressed in Section 5. 
\section{Dataset}
\begin{figure}[ht]
\begin{center}
\includegraphics[width=0.35\textwidth]{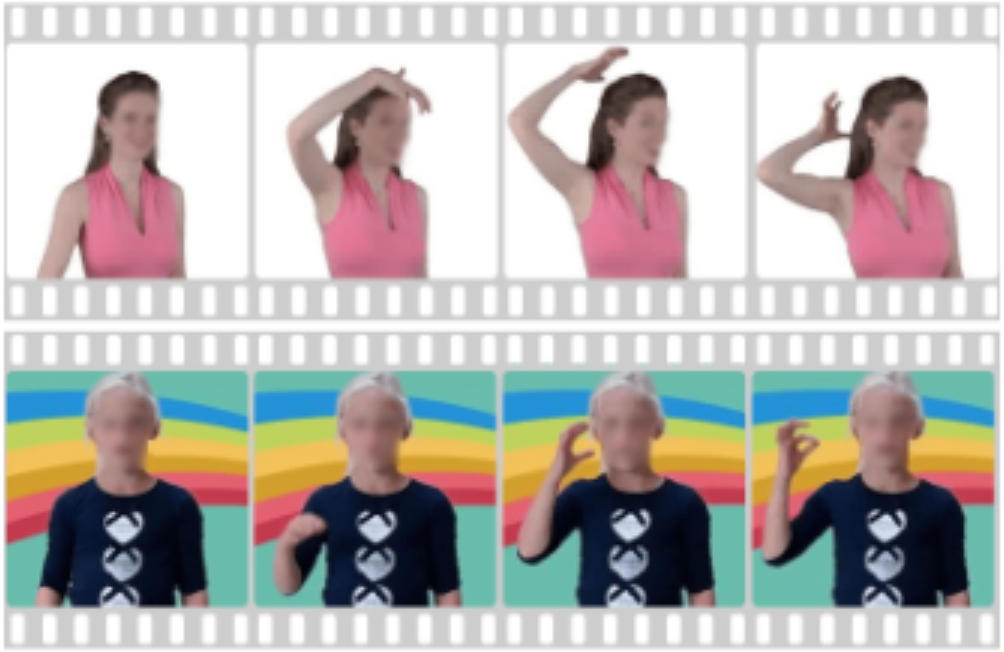}
\caption{ASL sequences of lion and cat signs
}
\label{fig:asl_samples}
\end{center}
\end{figure}

We collected our dataset via ASL dictionaries and resources on Internet for 10 different signs, particularly, referring to animals. These include bear, bird, cat, elephant, fish, giraffe, horse, lion, monkey and mouse. However, we exclude signs having more than one expression e.g. dog (this sign requires tapping on one's hip and optionally twisting thump and index fingers). Overall, we obtain 15 videos for each sign: 10 for training and 5 for testing. Our dataset presents a diversity of signers including 23 females, 10 males and 7 children. Each of them performs maximum 10 signs, minimum 1 sign (18 signers) and $3.75$ signs on average. In each video, we cut frames from starting of an action until it ends, within 1-2 seconds long  approximately. Figure \ref{fig:asl_samples} shows 2 signs sequences for lion and cat. We later extend the dataset to include 15 signs in Section 4.
\setlength{\textfloatsep}{0pt}
\section{Framework for Action Recognition}
\begin{figure}[ht]
\begin{center}
\includegraphics[width=0.45\textwidth]{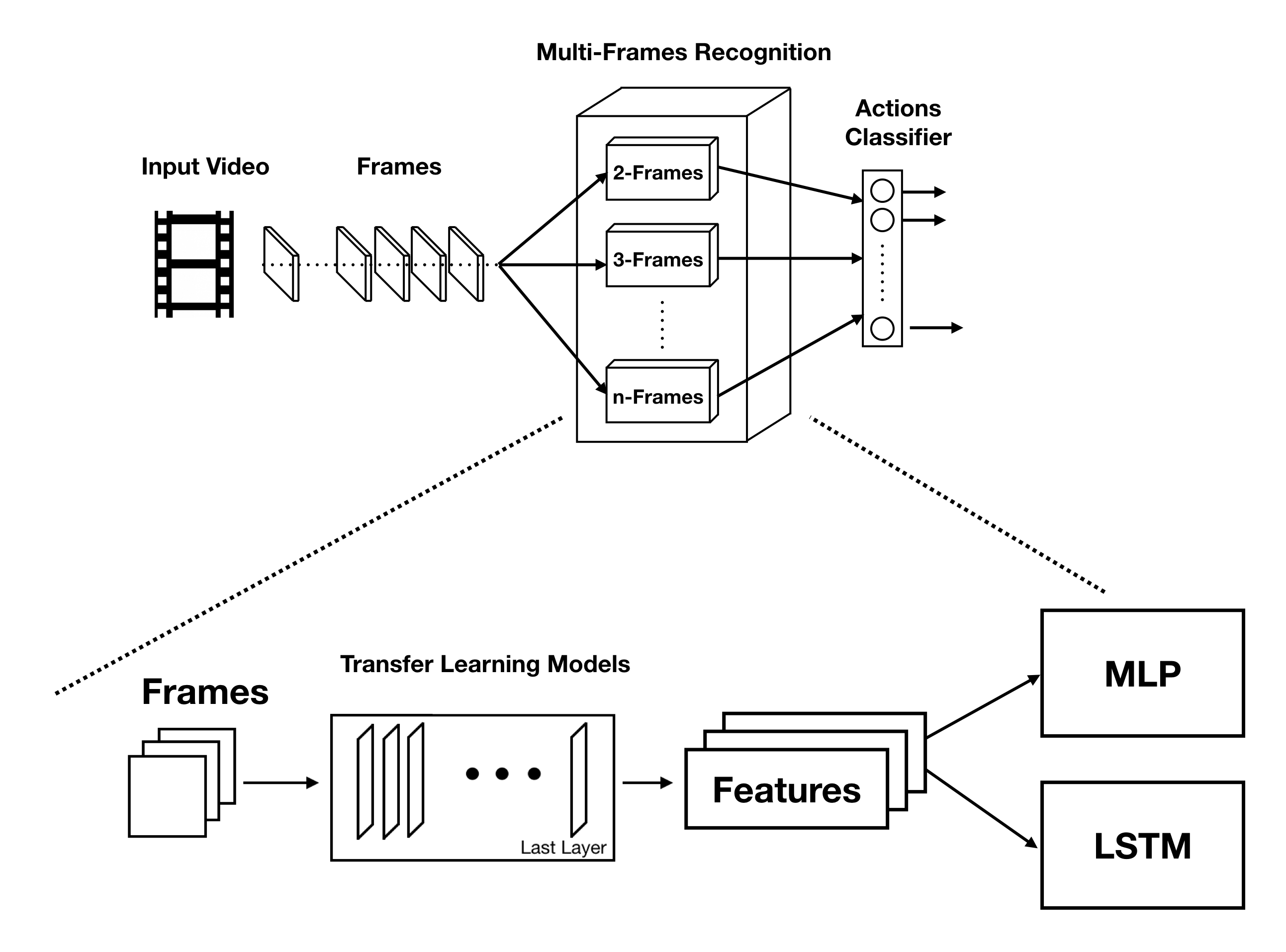}
\caption{Framework for Action Recognition}
\label{fig:action_recognition}
\end{center}
\end{figure}

We show our architecture for ASL Action Recognition in Figure~\ref{fig:action_recognition}. In this architecture, videos are extracted into frames at different rates e.g. 2 frames/s, 3 frames/s etc. Since visual contents usually share similar elements and training a descent deep net can take several weeks to months, for these reasons, we reuse trained deep networks to filter frames in our architecture and retrain the last layer for our dataset. At this step, we employ different transfer learning models naming InceptionV3, InceptionRestNetV2, Resnet50, DenseNet201  and VGG16/VGG19 to explore which model is the most suitable. Then these frames are extracted to a fixed set of features accordingly.

In classifying features from ASL signs, our approach is innovative compared to others in a similar problem endowing  classification of only one-shape signs even when signs are in continuous frames. Please note that an one-shape sign can be recognized using only one frame but a dynamic sign requires at least 2 or more frames. We apply Multi-layer Perceptron (MLP) as a baseline on each set of features since the neural network could perform comparable with other more advanced models \cite{phong2017offline}. Then we perform comparisons of MLP with Long Short Term Memory (LSTM). Performances of these structures are analyzed in next section.
\setlength{\textfloatsep}{0pt}
\section{Experiments \& Results}
In this section, we fist perform our experiments on ASL raw data for training phase and a pre-processing stage for the testing phase. We vary the length of sequence including 2, 4, 12 and 24 frames per video.

Table \ref{tab:experiment1} shows results for only InceptionResNetV2, InceptionV3 and DenseNet201. The results of ResNes50 and VGG16/VGG19 are excluded from this table since the accuracy is not much better than randomly guess. Performances of MLP and LSTM are also compared. We can observe that MLP usually outperforms LSTM despite of being a simpler structure. This is also interesting to notice that the accuracy of DenseNet201 is better than other models while it is not the finest model on ImageNet dataset. Regarding the choice of sequence length, a sign can be recognized with the highest accuracy using a sequence of just 12 frames. On the other hand, with only the first frame and the middle frame of an ASL sign, an accuracy of approximated $0.8$ can be observed.

In Experiment 2, we compare performances of MLP and LSTM on testing data using transfer learning model DenseNet201. From results for raw ASL data, we can observe that the architecture is not performing well and accuracy are just around $0.3$. For this reason, we pre-proceed background subtraction for videos and we also cut the first frame to remove backgrounds left by the subtraction process as we set history window to 1. In addition, the threshold is optimized to 50 since it gives the best accuracy. Beside, we use a median blur filter to remove noise in the frames. We can see from the result that the accuracy is much better with an improved accuracy of $0.58$ using LSTM on 12 frames per gesture.
%
\begin{table}[htb]
\centering
\caption{\label{tab:experiment1}ASL data classification on training DL models}
\begin{tabular}{lllllll}
             & \multicolumn{2}{c}{InceptionResNetV2} & \multicolumn{2}{c}{InceptionV3} & \multicolumn{2}{c}{DenseNet201}  \\ 
\cline{2-7}
             &        &                              &        &                        &        &                         \\
\#Seq Length & MLP    & LSTM                         & MLP    & LSTM                   & MLP    & LSTM                    \\ 
\hline
$2$          & -      & -                            & $0.86$ & $0.80$                 & $0.86$ & $0.88$                  \\
$4$          & -      & -                            & $0.97$ & $0.89$                 & $0.98$ & $0.89$                  \\
$12$         & -      & -                            & $0.96$ & $0.92$                 & $1.00$ & $0.91$                  \\
$24$         & $0.86$ & $0.75$                       & $0.93$ & $0.92$                 & $0.98$ & $0.94$                 
\end{tabular}
\end{table}

\begin{figure}[ht]
\begin{center}
\includegraphics[width=0.40\textwidth]{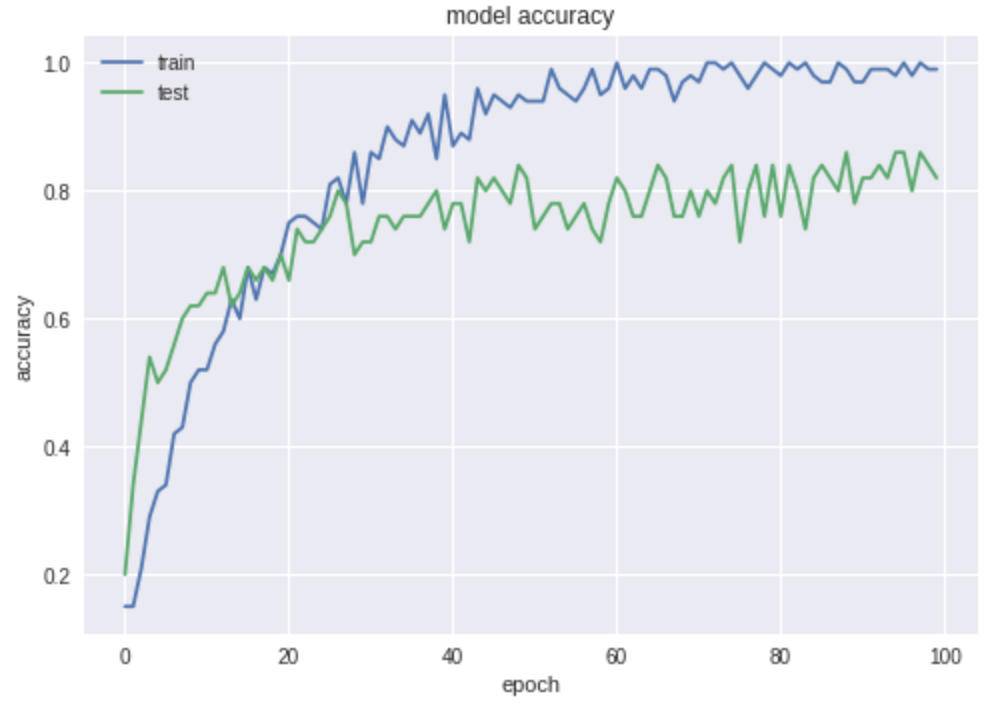}
\caption{Model Accuracy on DenseNet201 after Preprocessing}
\label{fig:animal_sign_accuracy}
\end{center}
\end{figure}

\begin{figure}[ht]
\begin{center}
\includegraphics[width=0.42\textwidth]{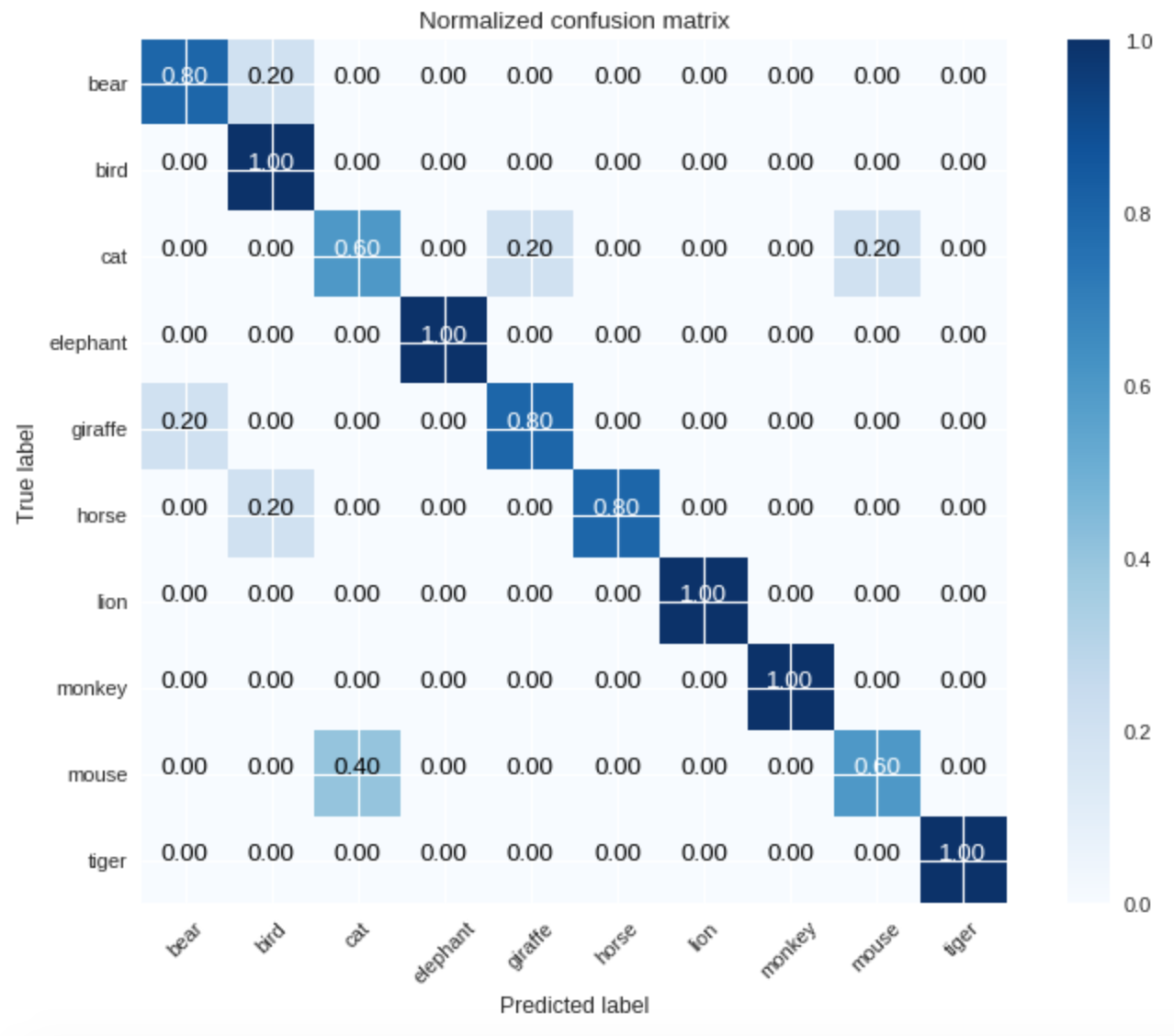}
\caption{Normalized Confusion Matrix with Accuracy of $0.86$ for 10 Signs}
\label{fig:animal_sign_normalized}
\end{center}
\end{figure}

We find out that the accuracy of $0.58$ may not be helpful for many applications and this could not getting better with the current setting. Further investigations point out that actors perform signs differently e.g. actors may repeat a gesture more often than others. For this reason, we strictly reinforce rules for these signs. 
Likewise,  initial signs were replaced for those we could not find enough videos according to the defined rules. After testing several times, we found that the combination of DenseNet201 and LSTM performs the best. Figure~\ref{fig:animal_sign_accuracy} and Figure~\ref{fig:animal_sign_normalized} show the accuracy and normalized confusion matrix for the training and testing. We observe the highest accuracy of $0.86$ yielded in test. However, in the confusion matrix, we notice that the model misclassifies between a mouse and a cat since their hand positions and shapes are somewhat similar.

\begin{figure}[ht]
\begin{center}
\includegraphics[width=0.42\textwidth]{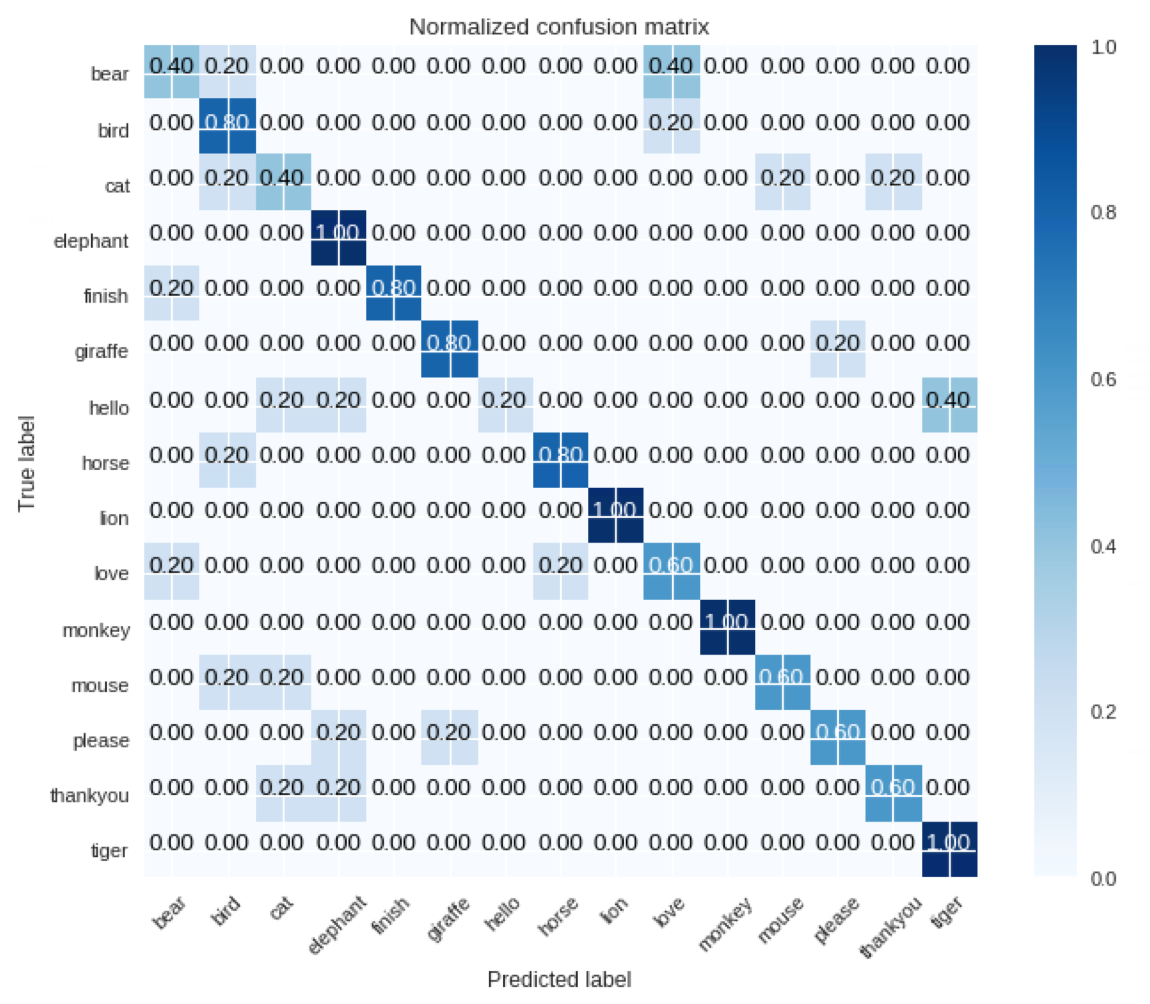}
\caption{Normalized Confusion Matrix for Extended Signs}
\label{fig:extend_sign_normalized}
\end{center}
\end{figure}

In Experiment 4, we extend our dataset from 10 signs to 15 signs including more general conversation terms as finish, hello, love, please and thank you. Confusion matrix is shown in Figure~\ref{fig:extend_sign_normalized}. We can see that while several classes e.g. elephant, lion, monkey and tiger are still being recognized correctly, other classes e.g. bear and love are mixed because of their similarity.
\setlength{\textfloatsep}{0pt}

\setlength{\textfloatsep}{0pt}

\setlength{\textfloatsep}{0pt}
\section{Conclusions and Future works}
In our main contribution, we explore the use of several transfer learning models in combination with deep neural networks to classify dynamic ASL signs. We found that an integration of DenseNet201 and LSTM performs the best. In addition, we collected our own database of 150 videos represent 10 signs and an extension of 225 videos that represent 15 signs for verification. We also vary the number of frames per sign to find the best setting.

Our results show that when we vary number of frames per sign, the use of 12 frames gives the highest accuracy. With two frames for one sign, the sign can also be recognized with accuracy of around $0.8$. On testing data, to obtain a higher accuracy, we subtract video backgrounds and strictly follow rules for signs. As a consequence, a highest accuracy of $0.86$ can be obtained. Our approach different from most of other researches which focus only on static ASL signs and another recent research \cite{ye2018recognizing} needs more than one input channel to obtain an accuracy of $0.69$.
Future work will address extending the dataset and improving our framework to better recognize signs and perform in realtime.

\bibliography{arfasl}

\begin{thebibliography}{5}
\providecommand{\natexlab}[1]{#1}
\providecommand{\url}[1]{\texttt{#1}}
\expandafter\ifx\csname urlstyle\endcsname\relax
  \providecommand{\doi}[1]{doi: #1}\else
  \providecommand{\doi}{doi: \begingroup \urlstyle{rm}\Url}\fi

\bibitem[Ameen and Vadera(2017)]{ameen2017convolutional}
Salem Ameen and Sunil Vadera.
\newblock A convolutional neural network to classify american sign language
  fingerspelling from depth and colour images.
\newblock \emph{Expert Systems}, 34\penalty0 (3):\penalty0 e12197, 2017.

\bibitem[Koller et~al.(2016)Koller, Ney, and Bowden]{koller2016deep}
Oscar Koller, Hermann Ney, and Richard Bowden.
\newblock Deep hand: How to train a cnn on 1 million hand images when your data
  is continuous and weakly labelled.
\newblock In \emph{Proceedings of the IEEE Conference on Computer Vision and
  Pattern Recognition}, pages 3793--3802, 2016.

\bibitem[Kuznetsova et~al.(2013)Kuznetsova, Leal-Taix{\'e}, and
  Rosenhahn]{kuznetsova2013real}
Alina Kuznetsova, Laura Leal-Taix{\'e}, and Bodo Rosenhahn.
\newblock Real-time sign language recognition using a consumer depth camera.
\newblock In \emph{Proceedings of the IEEE International Conference on Computer
  Vision Workshops}, pages 83--90, 2013.

\bibitem[Phong and Ribeiro(2017)]{phong2017offline}
Nguyen~Huu Phong and Bernardete Ribeiro.
\newblock Offline and online deep learning for image recognition.
\newblock In \emph{Experiment@ International Conference (exp. at'17), 2017
  4th}, pages 171--175. IEEE, 2017.

\bibitem[Ye et~al.(2018)Ye, Tian, Huenerfauth, Liu, Ruiz, Chong, Rehg, Palsson,
  Agustsson, Timofte, et~al.]{ye2018recognizing}
Yuancheng Ye, Yingli Tian, Matt Huenerfauth, Jingya Liu, Nataniel Ruiz, Eunji
  Chong, James~M Rehg, Sveinn Palsson, Eirikur Agustsson, Radu Timofte, et~al.
\newblock Recognizing american sign language gestures from within continuous
  videos.
\newblock In \emph{Proceedings of the IEEE Conference on Computer Vision and
  Pattern Recognition Workshops}, pages 2064--2073, 2018.

\end{thebibliography}
\end{document}